\documentclass[a4paper, conference]{IEEEtran}
\usepackage{amsmath}
\usepackage{graphicx}
\usepackage{xcolor}
\usepackage{hyperref}
\graphicspath{{./figures/}}

\begin{document}
\title{Improving Robotic Grasping on Monocular Images Via Multi-Task Learning and Positional Loss}

\author{\IEEEauthorblockN{William Prew\IEEEauthorrefmark{1}\IEEEauthorrefmark{3}, Toby Breckon\IEEEauthorrefmark{1}\IEEEauthorrefmark{2}, Magnus Bordewich\IEEEauthorrefmark{1}, Ulrik Beierholm\IEEEauthorrefmark{3}}
\IEEEauthorblockA{Department of {Computer Science\IEEEauthorrefmark{1}, Engineering\IEEEauthorrefmark{2}, Psychology\IEEEauthorrefmark{3}} \\
University of Durham, UK}}
\maketitle

\begin{abstract}
In this paper we introduce two methods of improving real-time object grasping performance from monocular colour images in an end-to-end CNN architecture. The first is the addition of an auxiliary task during model training (multi-task learning). Our multi-task CNN model improves grasping performance from a baseline average of 72.04\% to 78.14\% on the large Jacquard grasping dataset when performing a supplementary depth reconstruction task. The second is introducing a positional loss function that emphasises loss per pixel for secondary parameters (gripper angle and width) only on points of an object where a successful grasp can take place. This increases performance from a baseline average of 72.04\% to 78.92\% as well as reducing the number of training epochs required. These methods can be also performed in tandem resulting in a further performance increase to 79.12\%, while maintaining sufficient inference speed to afford real-time grasp processing.
\end{abstract}

\section{Introduction}
Robotic object grasping is a complex task, with solutions often drawing from the combined knowledge of diverse fields including physics, psychology, and computer science \cite{Gaussier2017, Caldera2018}.

Recently, machine learning has played a key role in simplifying the formation of grasp plans by reducing the need for hand-crafted analytical features specific to the task. For general-purpose robotics systems these hand crafted features can be difficult and time consuming to produce \cite{Caldera2018}. Machine learning offers  a more hands-off data-driven approach. Deep neural networks have successfully enabled robots to perform grasping tasks autonomously, with grasp performance often exceeding that of systems with human-designed features \cite{Bohg2014, Pinto2016a}, and with greater generalisability for unseen objects because new objects do not require unique analytical features.

Many architectures  of deep neural networks have been applied to the task of grasping. Previous works have utilised convolutional neural networks (CNN) \cite{Redmon2015, Kumra2017, Mahler2019, Morrison2018}, reinforcement learning systems (RL) \cite{Amarjyoti2017, Zeng2018, Kalashnikov2018}, and generative adversarial networks (GAN) \cite{Rao2020} in both real settings and simulated environments \cite{James2019} to form grasp plans.

\begin{figure}[t]
    \centering
    \includegraphics[width=\columnwidth]{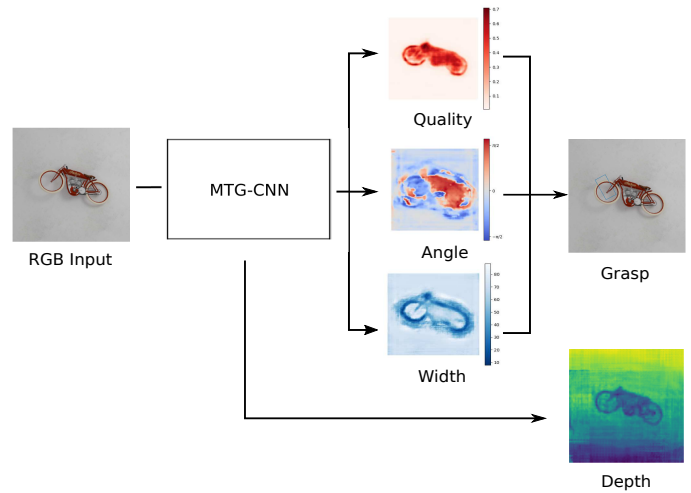}
    \caption{An overview of the MTG-CNN approach for grasp detection and depth reconstruction.}
    \label{fig:Overview}
\end{figure}

Most state-of-the-art systems tend to use complex network architectures, solving a number of simultaneous tasks such as semantic relations \cite{Ardon2019} which are combined to generate a grasp proposal. These networks therefore incur significant computational overhead to the detriment of real-time performance, often taking seconds to infer grasps \cite{Lenz2015, Mahler2017}. Smaller CNN architectures, while generally slightly less accurate, tend to run a lot faster and are capable of inferring grasps in real time \cite{Levine2018, Asif2019b}.

This paper proposes two methods of improving the accuracy of a lightweight CNN for robotic grasping while retaining real time performance: 1) incorporating multi-task learning during training, and 2) using a novel positional loss function. We build on the generative grasping convolutional neural network (GG-CNN) from Morrison et al. \cite{Morrison2018, Morrison2020}, a relatively small neural network which directly generates a grasp from an image. Due to the small number of parameters in the network, grasps can be generated multiple times per second for fast control of an autonomous robotic agent.

1) Multi-task learning, in which a system simultaneously learns to perform a main and an auxiliary task, has been shown to improve main task performance \cite{Caruana1998, Baxter2000} and has been increasingly applied to the field of machine learning to increase performance on a given task \cite{Ruder2017}. However, this is only seen when a network learns tasks that share a common optimal hypothesis task \cite{Ben-David2003}. We observe that grasping networks given true depth as an input can outperform networks given colour monocular images, and therefore propose our  multi-task grasping convolutional neural network (MTG-CNN). The MTG-CNN (see Fig. \ref{fig:Overview}) network replicates the GG-CNN network but also performs an auxiliary task. When this auxiliary task is to reconstruct depth from monocular images, we obtain improved performance at grasping without adding to the complexity of network architecture in deployment and hence preserve the ability for real-time grasp operation.

2) We introduce a new positional loss function which is able to achieve faster convergence, therefore requires far fewer training examples, and achieves better grasping performance on the large Jacquard grasping dataset \cite{Depierre2018}. The intuition behind the positional loss function is that we do not penalise the network for errors on the angle or width of grasp reconstructions where the position of the grasp is anyway unsuccessful, thereby focusing the learning of the network towards task success.

\section{Related Work}
The task of robotic grasping is defined as the technique in which a robotic end-effector can be positioned to securely grab an object and successfully lift it \cite{Caldera2018}. The action of planning and executing a grasp can be broken down into two parts: a computer vision, and a robotics problem.

For a successful vision based grasp plan, multiple factors should be considered. The proposed grasp should be robust enough to cope between variations in the environment, such as changes to lighting or viewing angle, and the inference time of a given model should also be small enough to provide a robotics system with a grasp plan in real time. Subsequently, the robotics system must be able to decide on the correct 3-D approach vector to achieve successful gripper closure on an object \cite{Bohg2014}. Ideally, these systems should complement one another to generalise to unseen objects.

In this section we briefly outline previous literature that relates to the formation of a successful grasp plan:

\subsection{Grasp Synthesis}
In the context of robotics, grasp synthesis refers to the problem of finding a grasp configuration that satisfies a set of criteria relevant for a grasping task \cite{Bohg2014}. According to Sahbani and colleagues \cite{Sahbani2012}, previous solutions can be broadly categorised into either analytical and empirical techniques.

Analytical techniques require the use of hand-crafted features, typically generated by humans, to generate application specific parameters. These hand-crafted features can include: specific geometric features, or mathematical or physical models to calculate stable grasps \cite{Bohg2014}. However, these are typically generated on the premise that these properties are known. Therefore it is difficult to generate such features for every object leading to problems with the generalisation of such techniques \cite{Bicchi2000}. 

Empirical (data-driven) techniques on the other hand, are developed using existing knowledge of object grasping and rely on previously known successful results \cite{Bohg2014}. One example of these has been the more recently applied deep learning models which have been used to generate grasp points via learned features such as grasp affordances \cite{Ardon2019} or semantic labels \cite{Xiang2018}. Machine learning models are now often used to form effective grasp plans for robots \cite{Mahler2017, Levine2018} including in real world settings e.g. \cite{Zhihong2017}. 

These empirical approaches are now the generally preferred method over their analytical counterparts due to their ability to generalise better to unseen objects \cite{Caldera2018} and their superior inference time allows for the real-time control of robotic agents \cite{Morrison2018, Asif2019b, Kumra2020}. Some models have even attempted to improve overall execution time by pre-processing grasp candidates \cite{Lenz2015} or trading off execution time for the number of sampled grasps \cite{Pinto2016a}. 

The main limitation of such empirical methods are that they require a large amount of detailed data in order to infer grasps \cite{Levine2018}. Some studies have attempted to improve generalisation to new environments with potentially far less available data with the introduction of techniques such as online learning using pretrained models \cite{Julian2020} or from transferring from simulation data to real world settings \cite{James2019}. However, these models tend to rely on highly detailed point cloud data \cite{Mahler2018} or colour and depth (RGB-D) input \cite{Kumra2017, Kumra2020} to train models which is not always available to certain systems. There is a section of research that attempts to improve results on impoverished colour only data by joint training on related tasks, inspired by mammalian models of vision \cite{Jang2017}. 

Here we build on the empirical work of Morrison et al. \cite{Morrison2018} by using a relatively small CNN. We are able to produce grasps with fast inference times by inferring grasp plans for every pixel simultaneously on a typical colour or depth image. The output from such models can then be passed to a robotic system for online, real-time control \cite{Morrison2020}.

\subsection{Grasping Datasets}
\begin{figure*}[ht]
    \centering
    \includegraphics[width=0.9\textwidth]{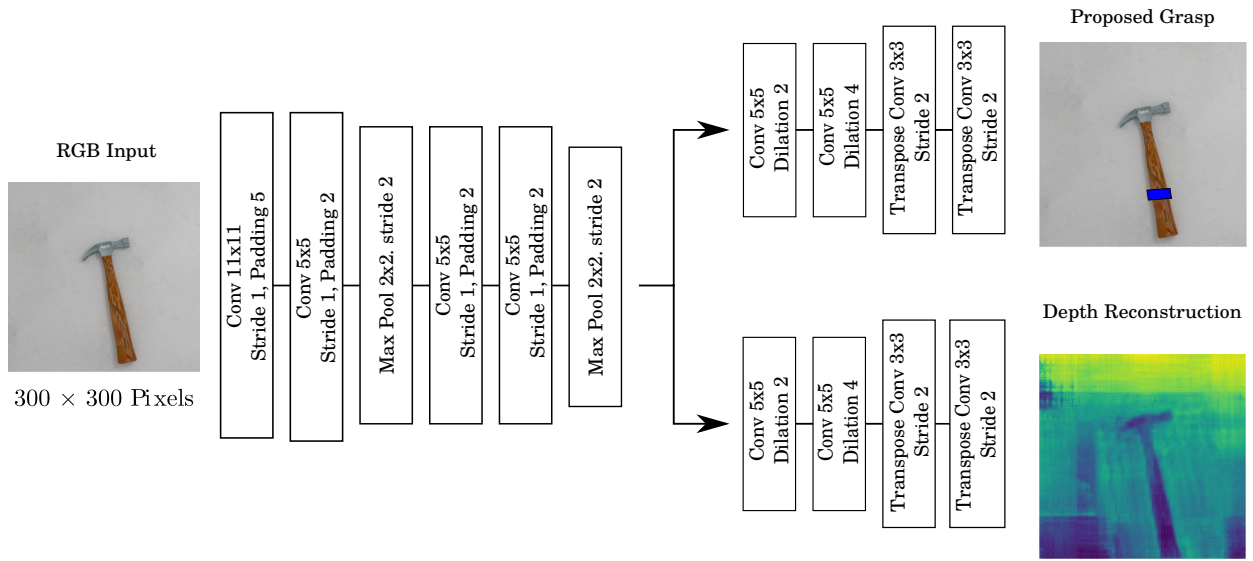}
    \caption{Architecture of the MTG-CNN network. The GG-CNN architecture follows the same structure except with the auxiliary task branch omitted.}
    \label{fig:network}
\end{figure*}

With the advent of data-driven grasping, new datasets have subsequently been proposed to provide large scale data reserves for training models. One of the earliest datasets was the popular and widely used Cornell grasping dataset (CGD) \cite{Saxena}. This contained an initially limited 885 images of 240 different object types, and a total of 8019 valid and invalid hand-labelled grasps. It was designed for common parallel plate grippers and was used extensively as a benchmark c.f. \cite{Kumra2017, Guo2017, Asif2019b}. As each grasp was labelled by a human annotator, Mahler \cite{Mahler2017} notes multiple drawbacks to the process, as it was tedious to collect all available grasps and therefore difficult to generalise the hand-labelling process to other objects or scenes. Furthermore, this process likely introduced human bias into the robotic grasping data as the annotator is likely to label a grasp that is easier for humans to pick up but not necessarily for a parallel plate gripper attached to a robot arm \cite{Pinto2016a, Depierre2018}.

Since the CGD, more contemporary datasets have been developed that aim to facilitate the formation of grasp synthesis plans on different robotic arms. Thereby, improving their ability to generalise to different environments. Some datasets have focused on generating a larger amount of data for empirical methods to train from \cite{Pinto2016a, Depierre2018}, or developing an open-source platform to consolidate data from multiple robotic arms \cite{Dasari2019}. Other datasets have focused on improving the replicability of previously developed models between different arms by producing datasets filled with standardised items \cite{Yang2019, morrison2020egad}, or developing more detailed datasets for grasping on robot arms with access to a greater number of degrees of freedom \cite{Xiang2018, Fang2020}. 

Most of these datasets have the common benefit over the CGD in that they have generated a larger amount of labelled grasps to train data-driven models. Another commonality is that most of these datasets are typically now labelled by the robot arms themselves compared to the hand-labelled data from the CGD for more accurate training data. This has been achieved through procedures such as autonomous collection from real robots \cite{Levine2018}, or by generating a large number of grasping examples using simulated data \cite{James2019}. This is done in both an online fashion through reinforcement learning \cite{Yang2019}, or by using a simulated robot arm to attempt a larger number of grasps like the more recent Jacquard grasping dataset \cite{Depierre2018} that was used to train and evaluate the model presented here. 


This is a large-scale grasping dataset where all grasps are developed and labelled in a simulated environment. It contains over 1.1 million grasps for over 11,000 objects which is a widely used benchmark amongst data-driven antipodal grasping models c.f. \cite{Morrison2018, Kumra2020}.

\subsection{Multi-Task Learning}


Multi-task learning (MTL) is a machine learning approach that aims to help models better generalise to a given main task by using the training signals of a related task to learn an appropriate inductive bias \cite{Caruana1998}. MTL has been shown to improve performance on a main task when simultaneously training on a simpler auxiliary task across a wide range of disciplines \cite{Ruder2017}. By sharing features between multiple tasks, the network is forced to learn common representations between them that may reduce overfitting, resulting in better generalisation to the original task. Examples include computer vision (e.g. using the characteristics of the road to predict steering direction \cite{Caruana1998} or the segmentation of images for classification in the fast R-CNN model\cite{Girshick2015}); natural language processing \cite{Collobert2008, Liu2020}; and speech recognition \cite{Deng2013}.

MTL can be implemented through an assortment of different functions, either separately or together in the same model; as long as the model has access to all the common representations required for both tasks. Such examples can include the cross-talk of information between different networks \cite{Liu2020}, focusing attention on a given task by learning an auxilliary training bias \cite{Caruana1998}, or leveraging more specific examples like representation learning for language modelling \cite{Rei2017}. Some studies do state for MTL to be effective that related tasks need to share a similar optimal inductive bias \cite{Baxter2000}. As such, it is not always clear which auxiliary tasks will lead to performance increases on the main task during joint learning \cite{Ruder2017}. Some studies have attempted to bypass this problem with specifically designed multi-task losses that add an extra coefficient for training a plethora of different techniques \cite{Cipolla2018}.

Most current deep learning implementations for robotic grasping tend to only focus on a single grasping task \cite{Redmon2015, Levine2018, Asif2019b}. However, specific grasping examples have shown that there is a relative performance increase when networks are trained while performing auxiliary tasks, such as semantic segmentation \cite{Katz2014, Dang2014, Jang2017, Xiang2018, Ardon2019}; saliency detection \cite{Klein2017}; or the generation of bounding boxes \cite{Nguyen2017}. Such MTL techniques can be applied to the model using either a bottom-up \cite{Klein2017, Ardon2019} or a top-down approach \cite{Jang2017, Xiang2018}. Similarly, in the current study we attempt to improve overall grasping performance by forcing shared parts of the network to learn a similar auxilliary task, when presented with only monocular colour data.

\section{Methodology}
In this section we give a formal definition of the grasping problem we address, then present the network architecture and loss functions used in our solution and finally describe the dataset used for training and evaluation. 

\subsection{Grasping Problem}
The grasping problem we address is to take a colour (RGB) input image $I$ of an object and output a proposed grasp for a robotic arm with an antipodal gripper approaching perpendicularly to the $x$-$y$ plane. The proposed grasp is given as a tuple $((x,y),\Phi,W)$, where $(x,y)$ is the central position of the gripper in the $x$-$y$ plane relative to the image, $\Phi$ is the angle of rotation of the gripper around the $z$-axis, and $W$ is the width of the gripper jaws for the grasp attempt.

A proposed grasp corresponds to real world coordinates through a series of known transforms in Cartesian coordinates via the below equation:
\begin{equation}
    ((x,y),\Phi,W) = t_{RC}(t_{CI}((x,y),\Phi,W))
\end{equation}
where $t_{CI}$ is the transform from the 2D image coordinates into the 3D camera frame, and $t_{RC}$ is the transform from the camera frame to the world or robot frame with known intrinsic parameters and known calibration between the camera and the robot.

\subsection{Multi-task Network Architecture} \label{Architecture}
Our proposed solution is a multi-task grasping convolutional neural network (MTG-CNN), based on the generative grasping convolutional neural network (GG-CNN) \cite{Morrison2018, Morrison2020}. The GG-CNN is a small and lightweight network for closed-loop continuous grasping to produce a map of potential antipodal grasps perpendicular to the $x$-$y$ plane. Our experiments use the GG-CNN model as defined in \cite{Morrison2020} as the backbone. 
The MTG-CNN network extends the GG-CNN network by branching at the halfway point of the GG-CNN network and duplicating the second half. One branch is used for learning the grasp outputs, as in GG-CNN, the other branch for an auxiliary learning task, shown as depth reconstruction in Fig.~\ref{fig:network}. 
The shared first half consists of two convolutional layers, followed by a max pooling layer and two further convolutional layers and a second max pooling layer. The two branches in the second half of the network are identical and consist of two dilation layers and two convolutional transpose layers, see Fig. \ref{fig:network}. All convolutional layers have 16 filters except for the dilated convolutional layers which have 32 filters.

\begin{figure}
    \centering
    \includegraphics[width=\columnwidth]{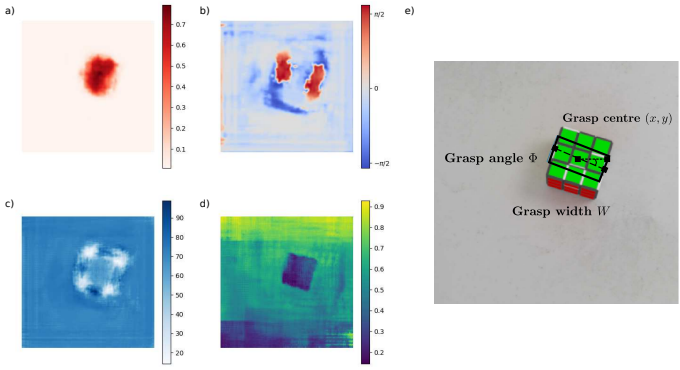}
    \caption{Example output from the MTG-CNN model that contributes towards a labelled grasp image. a) The probability of grasp success $Q$ for each pixel in the range of $[0,1]$, b) The proposed grasp angle $\Phi$ at each pixel generated during post-processing from $\Phi_{\sin}$ and $\Phi_{\cos}$ in the range $[-\frac{\pi}{2}, \frac{\pi}{2}]$. c) The required width of the gripper $W$ represented pixel-wise in the range $[0,150]$. d) The generated depth map image from an RGB input normalised between the values $[0,1]$. This does not contribute to the final grasp. e) An example grasp. The pixel with the highest grasp probability in Q is used as the centre of the grasping rectangle. The angle and width are calculated from the same pixel coordinates $(x,y)$ from each image.}
    \label{fig:output}
\end{figure}

The MTG-CNN network can be viewed as a function $M_\theta$, where $\theta$ represents the model weights, which maps a 300$\times$300 pixel input image $I$ to five parameter-space output images $Q_\theta(I), \Phi_{\theta}^{\sin}(I),  \Phi_{\theta}^{\cos}(I),W_\theta(I)$ and $A_\theta(I)$. The first four of these form the grasping output and come from one branch of the network, and the final auxiliary output $A_\theta$ comes from the the other, auxiliary, branch. The pixel values of $Q_\theta(I)$ represent the probability of a successful grasp at each pixel position in $I$. The angle of the proposed grasp can be calculated pixelwise by $\Phi_\theta(I) = \arctan(\frac{\Phi^{\sin}_{\theta}(I)}{\Phi^{\cos}_{\theta}(I)})/2$. 
The raw angle $\Phi$ of each grasp is in the range $[-\frac{\pi}{2}, \frac{\pi}{2}]$; it is represented as two unit vector components $\cos(2\Phi)$ and $\sin(2\Phi)$ in the network outputs as this aids learning by removing the discontinuity at the wrap-around point~\cite{Hara2017}. 
Finally, $W_\theta(I)$ is the required gripper width as a value in the range $[0,1]$, which is scaled into a physical measurement for gripper width in the range of $[0,150]$. Fig.~\ref{fig:output} illustrates an example input and outputs of MTG-CNN.

The GG-CNN network, which does not feature the auxiliary branch, is similar but produces only the four parameter space output images $Q_\theta(I), \Phi_{\theta}^{\sin}(I),  \Phi_{\theta}^{\cos}(I)$ and $W_\theta(I)$ corresponding to the grasping outputs from the original branch of the MTG-CNN network.

To extract a proposed grasp from either network we take the $x,y$ coordinates of the maximum pixel value in $Q_\theta(I)$ as the centre of the grasp, and then extract the corresponding pixel values from $\Phi_{\theta}(I)$ and $W_\theta(I)$ to obtain the angle and width of the gripper.  For smoothness, before extracting the proposed grasp the outputs $Q_\theta(I)$, $W_\theta(I)$ and $\Phi_\theta(I)$ are filtered with  a Gaussian kernel and a standard deviation of 5 pixels, as is done in~\cite{Johns2016}. 

At inference (test) time only the parameterized output of the grasp is required, so the additional complexity of the auxiliary branch in MTG-CNN can be discarded resulting in streamlined real-time capable performance.

The purpose of multi-task learning is to guide the shared layers of the network towards learning features that will be useful for the grasping task, which it might not otherwise find. Preliminary experiments on the GG-CNN network given true depth images as input, instead of colour images, showed improved performance at grasping. This motivated us to use depth reconstruction as an auxiliary task, with the intuition that the MTG-CNN network would learn features related to 3D structure in the shared layers, and these would be useful for the grasping task. In addition to depth reconstruction, we also tested saliency detection as an auxiliary task, with the intuition that the network needs to identify the object in the scene, and features tuned towards this task in the shared layers may be helpful.

\subsection{Positional Loss}

This paper introduces a more focused loss function, which we call \textit{positional loss}, for use during training. The loss function used with the GG-CNN network in~\cite{Morrison2020} is the summation of the MSE loss of each of the four parameter space output images $Q_\theta(I), \Phi_{\theta}^{\sin}(I),  \Phi_{\theta}^{\cos}(I),W_\theta(I)$. This can be extended to a loss function $L$ for the MTG-CNN network by simply including the MSE of the auxiliary output in addition:

\begin{equation} \label{loss}
\left.\begin{aligned}
L {}={} & \frac{1}{N} \sum (Q_{\theta} - Q_{GT})^2\\
{}&+  \frac{1}{N} \sum (\Phi_{\theta}^{\cos} - \Phi^{\cos}_{GT})^2\\
{}&+  \frac{1}{N} \sum (\Phi_{\theta}^{\sin} - \Phi^{\sin}_{GT})^2\\
{}&+  \frac{1}{N} \sum (W_{\theta} - W_{GT})^2\\
{}&+  \frac{1}{N} \sum (A_{\theta} - A_{GT})^2\\
\end{aligned}\right.
\end{equation}%

where $N$ is the number of pixels per image, the summations are over the pixels of each image, $\theta$ denotes the output of the network and $GT$ denotes the ground truth from the dataset.

The loss function $L$ includes a significant contribution from the error in angle prediction and width prediction even at positions of the input image where no successful grasp is possible, for example: regions containing the background not the object. The network must learn to output values close to a default value of 0 for angle and width in such regions even though no grasp will be attempted at these positions and therefore these angle and width parameters are not used. The positional loss function $L_p$ scales the angle and width errors by ground truth for the quality of the grasp ($Q_{GT}$) at each position in the scene. The scaling of the loss in this manner focuses the learning of the network on grasp quality and on angles and widths only at positions where grasp attempts may occur. The positional loss function is defined below:

\begin{equation} \label{p_loss}
\left.\begin{aligned}
L_p {}={} & \frac{1}{N} \sum (Q_{\theta} - Q_{GT})^2\\
&{+}  \frac{1}{N} \sum (Q_{GT}(\Phi_{\theta}^{\cos} - \Phi^{\cos}_{GT}))^2\\
&{+}  \frac{1}{N} \sum (Q_{GT}(\Phi_{\theta}^{\sin} - \Phi^{\sin}_{GT}))^2\\
&{+}  \frac{1}{N} \sum (Q_{GT}(W_{\theta} - W_{GT}))^2\\
&{+}  \frac{1}{N} \sum (A_{\theta} - A_{GT})^2.\\
\end{aligned}\right.
\end{equation}

For the GG-CNN model, the loss functions $L$ and $L_p$ are as above with the final summation, with the extra auxiliary term removed. Networks that are trained with the new positional loss function are marked with a subscript $p$.

\subsection{Dataset} \label{Dataset}

All models are trained with the Jacquard grasping dataset \cite{Depierre2018}. This is a simulated dataset containing 54,485 images of over 11,000 different objects on uniform white backgrounds. The images are annotated with over 1.1 million successful grasps represented as grasp centre, angle and gripper width. All grasps were generated within a simulated physics environment which attempted grasps at many positions, angles and widths; unsuccessful grasps and highly similar grasps were not recorded.

Every object in the dataset has four viewing angles and each viewpoint consists of a single RGB image as well as a perfect depth image recorded from the simulated data and a generated stereo depth image. The perfect depth image was used to train the depth reconstruction branch. 

\begin{figure}[t!]
    \centering
    \includegraphics[width=\columnwidth]{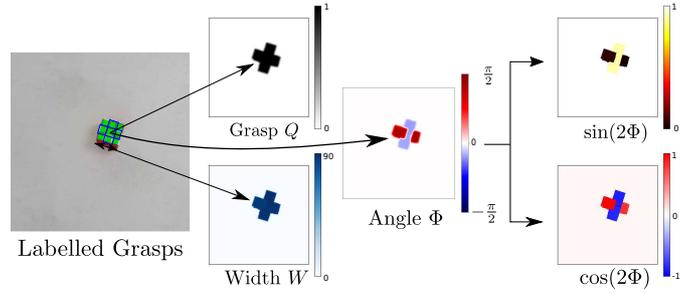}
    \caption{Annotated grasps from the dataset are transformed into images for training. Grasps are broken down into grasp quality $Q$, grasp angle $\Phi$, and grasp width $W$.}
    \label{fig:Dataset}
\end{figure}

For each image, $I$, ground truth parameter-space images, $Q_{GT}(I), \Phi_{GT}^{\sin}(I),  \Phi_{GT}^{\cos}(I),W_{GT}(I)$ and $A_{GT}(I)$, are computed as follows (see Fig.~\ref{fig:Dataset}). For $Q_{GT}(I)$ the pixel values are 1 if the pixel falls within the centre third of the successful grasping rectangle for $I$ and 0 otherwise. Values of $\Phi_{GT}^{\cos}(I),\Phi_{GT}^{\sin}(I)$ and $W_{GT}(I)$ are set to according to the angle and width of a corresponding successful grasp centred at that pixel position, where $Q_{GT}(I)=1$ and are set to 0 elsewhere. The angle $\Phi$ of the grasp is in the range $[-\frac{\pi}{2}, \frac{\pi}{2}]$, and is decomposed into two unit vector components $\cos(2\Phi)$ and $\sin(2\Phi)$, which are used for $\Phi_{GT}^{\cos}(I),\Phi_{GT}^{\sin}(I)$.
When the auxiliary branch is learning depth reconstruction, $A_{GT}(I)$ is set to the true depth
data of the simulated scenes normalised to the range $[0,1]$. When the auxiliary branch is learning saliency, $A_{GT}(I)$ is set to the value of a binary mask provided with the dataset, where 1 indicates a pixel displaying part of the object, and 0 a pixel displaying the background.

For training and testing, the dataset is split in a similar manner to~\cite{Depierre2018}: a 5-fold cross validation was performed in which the 80\% of the data was reserved for training, and
the remaining 20\% of the data was split evenly between a
validation and test set. Colour input data was normalised to the range of $[0,1]$ before subtracting the image mean to centre the data around 0. All models were trained for 40 epochs which preliminary testing showed was sufficient for full training. Furthermore, data augmentation was applied to the dataset to artificially increase the amount of training data by random cropping and zooming on the images.

Performance on the dataset is measured using the intersection-over-union (IoU): a grasp is considered successful if the predicted grasp rectangle shares a greater than 25\% IoU and aligns within $30^{\circ}$ with the ground truth grasping rectangle. This measure provides an efficient offline metric for testing performance and was used in both the Jacquard \cite{Depierre2018} and Cornell \cite{Saxena} datasets, the values here are from the Jacquard dataset \cite{Depierre2018}. 


\begin{figure}[t!]
    \centering
    \includegraphics[width=\columnwidth]{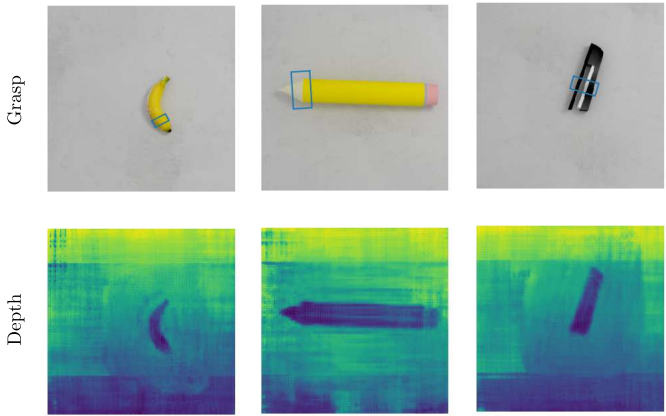}
    \caption{Example grasp candidates and depth reconstruction from the MTG-CNN network. We suspect that previous compression of images has caused a boxing artefact that results in a grid-like structure during depth reconstruction.}
    \label{fig:Output2}
\end{figure}

\section{Evaluation} \label{Evaluation}
\begin{figure*}
    \centering
    \includegraphics[width=\textwidth]{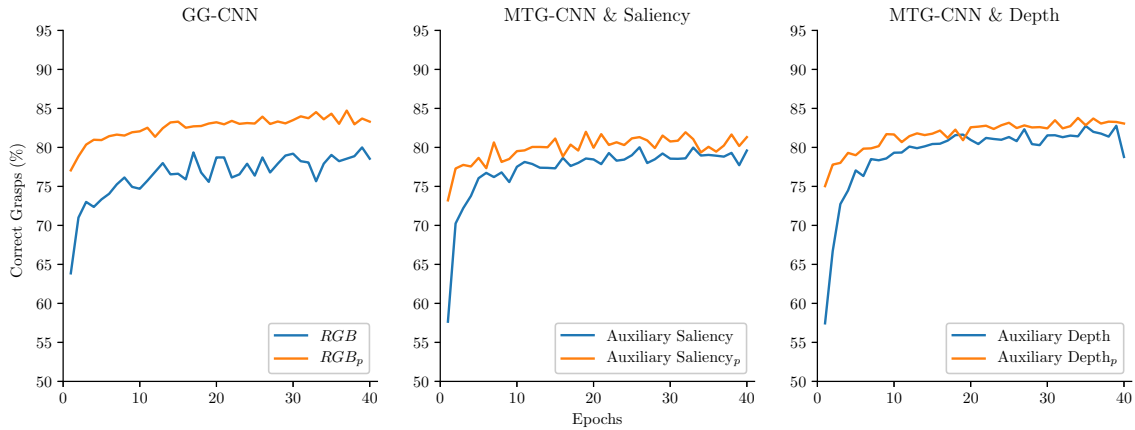}
    \caption{Average grasping performance of each model on the validation set after each epoch. Models trained with the new positional loss function maintain faster convergence than those trained with just the previous grasping loss.}
    \label{fig:validation}
\end{figure*}

A 5-fold cross validation was used to evaluate all the models. After training was completed, each network was tested on every object in the test set, and the percentage of successful grasp proposals under the IoU metric recorded (see Section~\ref{Dataset}). The average number of correct grasps and standard deviation of these across the five test folds is shown in Table \ref{Grasps}.

\begin{table}
\renewcommand{\arraystretch}{1.3}
\caption{The mean percentage of correct grasps for each network evaluated on the Jacquard grasping dataset.}
\label{Grasps}
\centering
\begin{tabular}{c c c}
\hline
\bfseries Model & \bfseries Auxiliary Task & \bfseries Successful Grasps \\
\hline
GG-CNN   & & $72.04\% \pm 3.44$\\
\hline
\hline
GG-CNN$_p$  & &$ 78.92\% \pm 0.97$\\
\hline
MTG-CNN  & Saliency & $74.93\% \pm 1.86$\\
MTG-CNN$_p$ & Saliency & $76.23\% \pm 2.75$\\
\hline
MTG-CNN  & Depth & $78.14\% \pm 0.65$\\
\textbf{MTG-CNN}$\boldsymbol{_p}$ & \textbf{Depth} & $\boldsymbol{79.12\% \pm 1.40}$\\
\hline
\end{tabular}
\end{table}

These results show that by training on either auxiliary task we tested, the MTG-CNN network performs better than the base GG-CNN model. With the auxiliary saliency detection task, the grasping performance increases from the average baseline of 72.04\% to 74.93\%. With the auxiliary depth reconstruction task we see a larger increase, an average performance of 78.14\%, and the network becomes more consistent overall. This implies that by training a deep neural network to perform simultaneous depth reconstruction, we can improve the results for the primary grasping task by forcing the early shared part of the network to learn key structural information about the object. We include some example outputs of the MTG-CNN network trained on depth in Fig. \ref{fig:Output2}.

It can also be seen that training with the positional loss function identifies a greater number of correct grasps compared to training with the previous GG-CNN loss function: grasping performance increases from an average of 72.04\% to 78.92\% on the GG-CNN network, and also shows an increase for the MTG-CNN network with either auxiliary function. 

When both an auxiliary task and the positional loss function are combined, we obtain the best results. The best performing network is the model trained on the auxiliary depth reconstruction task with 79.12\%. At first glance these results seem show that the positional loss function only provides a small additional benefit to performance for the MTG-CNN architecture, in particular when the auxiliary function is depth reconstruction. However, by looking at Fig. \ref{fig:validation} we see that all networks trained on the new positional loss function perform better on average that the previous loss function. During the initial parts of training, the positional loss function is achieving higher validation performance compared to the previous loss function. Therefore, the positional loss networks are actually learning the grasping task at a quicker rate, as well as attaining better performance with fewer training examples.


The GG-CCN model has 70,000 parameters, whereas the MTG-CNN model has 115,000 parameters, due to the additional branch. This slightly increases the training time per epoch, but the network remains significantly smaller than typical deep learning grasping networks. Furthermore, we do not increase inference times significantly. On a desktop NVIDIA GTX 1080Ti GPU-accelerated PC we achieve inference times between 10-25ms for the base GG-CNN model and 15-30ms for the dual-task MTG-CNN model, thus both models operate at over 30FPS. Moreover, during deployment the auxiliary task does not contribute to the grasp formation and therefore the auxiliary branch of the network can be removed. This results in identical inference times for MTG-CNN and GG-CNN, at at an average of over 50FPS.

It should also be noted that we chose the depth reconstruction as it has been shown in previous work that depth information provides key structural object information required for grasping \cite{Morrison2018}. It was believed that this task would improve the performance on monocular data according to studies like He et al. \cite{He2019} and Raffel et al. \cite{Raffel2019} that show similar tasks achieve greater performance which is the reasoning behind using MSE loss for both grasping and auxiliary tasks. However, by analysing our depth reconstructions from the model, this type of loss function appears to produce artefacts that we believe to be as a result of image compression during data collection.

In future, performance could be improved with tailored loss functions other than typical MSE loss. Depth specific reconstruction losses could be used that achieve a greater smoothing effect such as that of \cite{godard2017unsupervised} or adding the use of specific multi-task loss functions like \cite{Cipolla2018}. Alternatively, tasks that are not directly related to grasping like semantic understanding \cite{Xiang2018} or grasp affordances \cite{Ardon2019} may also be applicable and explored in follow up work. 

\section{Conclusion}
The first aim of this paper was to investigate whether multi-task learning can improve grasping performance when trained on auxiliary tasks with a relationship to grasping. By performing concurrent depth reconstruction with grasping, we significantly improve grasp performance on a lightweight network with only a small number of trainable parameters, and retain the speed of grasp inference.

Secondly, by introducing a new positional loss function we can train a model that converges faster and attains a further increase in grasp accuracy compared to the previous implemented loss function. The positional loss function removes learning of unnecessary output parameters (angle and gripper width) for output values of low grasp quality which do affect final grasp output, thereby allowing the network to focus learning on parameters affecting task performance itself. 

\bibliographystyle{IEEEtran}
%
\bibliography{main.bib}

\end{document}